\documentclass[a4paper]{cas-sc}

\usepackage{microtype}
\usepackage{graphicx}
\usepackage{subcaption}
\usepackage{booktabs} % for professional tables
\usepackage{array}
\usepackage{multirow}     % 支持 \multirow
\usepackage{makecell}
\usepackage[table]{xcolor}
\usepackage{bbding}
\usepackage{adjustbox}

\usepackage{hyperref}

\usepackage[authoryear]{natbib}
\def\tsc#1{\csdef{#1}{\textsc{\lowercase{#1}}\xspace}}
\tsc{WGM}
\tsc{QE}
\begin{document}
\let\WriteBookmarks\relax
\def\floatpagepagefraction{1}
\def\textpagefraction{.001}

\makeatletter
\def\printorcid{}
\def\orcidlink#1{}
\makeatother
\shorttitle{}    

\shortauthors{}  

\title [mode = title]{Cross-Modal Clinical Knowledge Integration for Mammography Report Generation}  




\author[1]{Jiayi Zhu}\fnmark[1]
\credit{Conceptualized and designed the study, assisted with data processing, including data cleaning and post-processing, designed and developed MammoRGTool and MammoRG, designed and conducted all experiments, and wrote the manuscript}
% \cormark[1]

% % Footnote of the first author

% % Email id of the first author
% \ead{}

% % URL of the first author

% % Credit authorship
% % eg: \credit{Conceptualization of this study, Methodology, Software}


\author[2]{Fuxiang Huang}\fnmark[1]
\credit{Assisted in conceptualizing and designing the study, and collected and processed the data}

\author[3]{Yu Xie}\fnmark[1]
\credit{Collected and annotated the datasets from a hospital, and provided clinical assessment recommendations}

\author[4]{Xi Wang}
\credit{Provided critical feedback on the manuscript and experimental design}
\author[4]{Zhixuan Chen}
\credit{Provided critical feedback on the manuscript and experimental design}
\author[5]{Yuan Guo}
\credit{Collected and annotated the datasets from a hospital}
\author[6]{Qingcong Kong}
\credit{Collected and annotated the datasets from a hospital}
\author[3]{Zhenhui Li}
\credit{Collected and annotated the datasets from a hospital}
\author[1,4]{Qiong Luo}\cormark[1]\ead{luo@cse.ust.hk}
\credit{Provided critical feedback on the manuscript and experimental design, and supervised the research}
\author[4,7]{Hao Chen}\cormark[1]\ead{jhc@ust.hk}
\credit{Provided critical feedback on the manuscript and experimental design, and supervised the research}
\affiliation[1]{organization={The Hong Kong University of Science and Technology (Guangzhou)},
            city={Guangzhou},
%          citysep={}, % Uncomment if no comma needed between city and postcode
            state={Guangdong},
            country={China}}
\affiliation[2]{organization={Lingnan University},
            % city={Hong Kong},
%          citysep={}, % Uncomment if no comma needed between city and postcode
            state={Hong Kong},
            country={China}}
\affiliation[3]{organization={The Third Affiliated Hospital of Kunming Medical University, Yunnan Cancer Hospital, Peking University Cancer Hospital Yunnan},
            city={Kunming},
%          citysep={}, % Uncomment if no comma needed between city and postcode
            state={Yunnan},
            country={China}}
\affiliation[4]{organization={The Hong Kong University of Science and Technology},
            % city={Kunming},
%          citysep={}, % Uncomment if no comma needed between city and postcode
            state={Hong Kong},
            country={China}}
\affiliation[5]{organization={Guangzhou First People's Hospital, South China University of Technology},
            city={Guangzhou},
%          citysep={}, % Uncomment if no comma needed between city and postcode
            state={Guangdong},
            country={China}}
\affiliation[6]{organization={The Third Affiliated Hospital, Sun Yat-Sen University},
            city={Guangzhou},
%          citysep={}, % Uncomment if no comma needed between city and postcode
            state={Guangdong},
            country={China}}
\affiliation[7]{organization={HKUST Shenzhen-Hong Kong Collaborative Innovation Research Institute},
            city={Shenzhen},
%          citysep={}, % Uncomment if no comma needed between city and postcode
            state={Guangdong},
            country={China}}

\cortext[1]{Corresponding author}

\fntext[1]{These authors contributed equally to this work.}


\begin{abstract}
Breast cancer is a major global health concern, and mammography screening plays a central role in early detection. The large volume of screening examinations creates a substantial workload for radiologists, making accurate and consistent report generation a critical clinical challenge. Existing automated mammography report generation methods primarily focus on direct visual-to-text mapping, while overlooking the structured clinical reasoning process followed by radiologists in real-world practice. To address this limitation, we propose MammoRG, a mammography report generation framework that explicitly simulates the clinical reporting workflow by following the BI-RADS guideline and incorporating prior clinical knowledge to produce diagnostic reports. Specifically, MammoRG adopts a two-stage training framework. In the first stage, the model learns to integrate clinically relevant prior knowledge from a patient’s four-view mammograms through classification-based supervision. In the second stage, a terminology-aware supervised fine-tuning strategy is introduced to model mammography-specific clinical terms as atomic semantic units, enabling the generation of high-quality reports with improved clinical consistency. To facilitate clinical efficacy evaluation of generated reports, we further develop MammoRGTool, a dedicated mammography report parsing tool that extracts structured clinical information from free-text reports. Extensive experiments demonstrate that MammoRG consistently outperforms existing methods across multiple clinical efficacy metrics, particularly in diagnosis-related BI-RADS F1, where it surpasses the second-best model by 2.73\%, 2.04\%, 1.90\%, and 3.27\% on the internal, external 1, external 2, and VinDr-Mammo datasets, respectively.
\end{abstract}

% Keywords

\maketitle

\section{Introduction}

Breast cancer is one of the most significant health challenges globally and a leading cause of cancer-related deaths among women, with millions of new cases diagnosed each year \citep{broeders2012impact,morra2015breast}. Mammography stands as the most effective screening method for the early detection of breast cancer, and its widespread application has been proven to significantly reduce mortality rates \citep{jalalian2013computer,mammography2019diagnosis}. With the widespread implementation of breast cancer screening worldwide, the number of images that radiologists need to interpret has increased dramatically, imposing a substantial workload \citep{abbey2020sequential}. Radiology report generation is the task of automatically producing free-text descriptions of radiological images, aiming to provide comprehensive and structured textual summaries for given patient scans. Due to its great potential in alleviating radiologists' workload, improving the efficiency of diagnostic workflows, and reducing human error, this field has attracted widespread attention in recent years, leading to the emergence of numerous report generation methods across various imaging modalities, including X‑ray \citep{zambrano2025clinically}, CT \citep{li2025towards}, MRI \citep{lei2025interpretable}, and histopathology \citep{tran2025generating}.

However, despite the rapid advancement of report generation technology, existing general-purpose radiology report generation models cannot be directly adapted to mammography. Most existing methods are primarily designed for generic single-image visual-to-text generation and often fail to model the mammography-specific diagnostic reasoning workflow required in clinical practice. In particular, mammography report generation is inherently a multi-view diagnostic task that requires joint reasoning across bilateral breasts and multiple standardized views, including right craniocaudal (R-CC), right mediolateral oblique (R-MLO), left craniocaudal (L-CC), and left mediolateral oblique (L-MLO) views, to accurately characterize fine-grained findings and ensure consistent diagnostic assessment. Even powerful general models such as GPT-5 \citep{singh2025openai} are unable to effectively address this challenge (Fig. \ref{fig:toy} (a)). Beyond describing findings, mammography reporting must further assess malignancy risk and assign appropriate Breast Imaging Reporting and Data System (BI-RADS) categories, which directly guide clinical decision-making \citep{liberman2002breast}.
Despite the clinical importance of mammography report generation, prior work remains limited, with few publicly reproducible models and benchmarks, hindering systematic validation and comparison \citep{yalunin2021generating,de2025convnext,cao2025mammovlm}. Moreover, existing approaches—including general-purpose report generation models—primarily focus on direct visual-to-text mapping and often underutilize mammography-specific clinical priors (Fig. \ref{fig:toy} (b)), such as BI-RADS–grounded assessment criteria and established reporting conventions. However, the clinical value of a mammography report depends not only on linguistic fluency, but more importantly on the correctness of clinically relevant findings and diagnostic assessments. Therefore, developing a robust and open-resourced mammography report generation framework that can effectively incorporate mammography-specific clinical knowledge is crucial for advancing this field toward clinical utility.

Inspired by these challenges, we propose MammoRG, a two-stage training framework for mammography report generation that leverages cross-modal clinical knowledge integration and term-aware supervised fine-tuning to improve both linguistic quality and clinical diagnostic accuracy. Specifically, the proposed cross-modal clinical knowledge integration module incorporates clinically relevant prior knowledge from a report database and a mammography knowledge graph into the report generation process (Fig. \ref{fig:toy} (c)), enabling better alignment between visual features and diagnostic semantic information. To address the inherent limitations of pretrained language models in tokenizing and semantically understanding professional terminology, we further expand the language model vocabulary with a mammography lexicon during the term-aware supervised fine-tuning stage. To facilitate clinical efficacy (CE) evaluation of generated reports, we additionally develop MammoRGTool, a dedicated mammography report parsing tool that extracts structured clinical information from free-text reports, including BI-RADS categories, tissue composition types, abnormal findings, and their associated anatomical locations, descriptive terms, and diagnostic suggestions. Experiments on an internal test set, two independent external test sets, and a publicly available external dataset demonstrate that MammoRG achieves state-of-the-art (SOTA) performance on both natural language generation (NLG) metrics and CE metrics. We summarize our contributions as follows:

\begin{figure}
\centering
\includegraphics[width=\columnwidth]{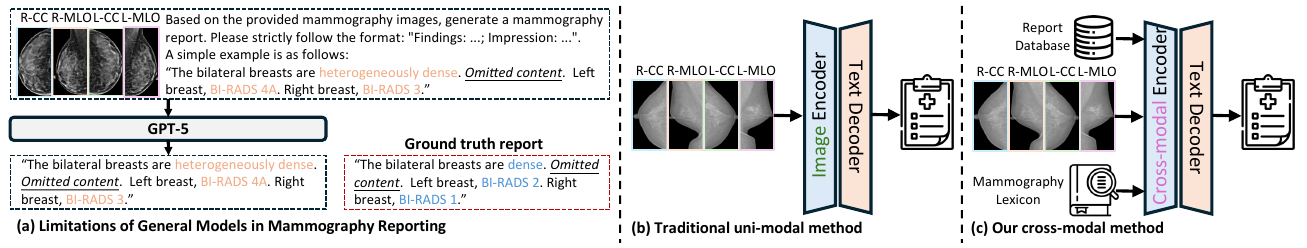}
\caption{Examples demonstrating limitations of general models in mammography reporting and the comparison between traditional uni-modal methods and our cross-modal method.}
\label{fig:toy}
% \vspace{-2em}
\end{figure}

\begin{itemize}
    \item We propose MammoRG, a two-stage training framework that leverages cross-modal clinical knowledge integration and term-aware supervised fine-tuning, effectively enhancing both the linguistic fluency and diagnostic accuracy of generated reports.
   
    \item We develop MammoRGTool, a specialized report parsing tool for evaluating mammography reports, which achieves a high macro-F1 score of over 99\% on an external validation set and establishes a reliable benchmark for the field.
   
    \item Extensive evaluation on the internal and three external test datasets demonstrates that MammoRG achieves SOTA performance on both NLG and CE metrics.
\end{itemize}

\section{Related Work}
\subsection{Radiology Report Generation}
Radiology report generation has evolved from early image-to-text frameworks toward multimodal reasoning systems that emphasize diagnostic accuracy and clinical reliability. Most existing methods are developed for chest X-ray or CT imaging, while mammography-specific report generation remains relatively underexplored~\citep{yalunin2021generating}.

Early radiology report generation methods mainly relied on encoder–decoder architectures to directly translate visual features into free-text reports~\citep{wang2020unifying,nishino2022factual,najdenkoska2022uncertainty}. More recent approaches incorporate pretrained language models and large language models to improve linguistic fluency and clinical coherence~\citep{devlin2019bert,moon2022multi,selivanov2023medical}. However, these methods are primarily designed for general radiology report generation and often treat report generation as a generic visual-to-text mapping problem. Unlike general radiology reporting, mammography report generation requires joint diagnostic reasoning across bilateral breasts, multiple standardized views, and BI-RADS assessment criteria to produce clinically meaningful diagnostic conclusions. Consequently, existing paradigms often struggle to model the mammography-specific clinical reasoning workflow.

To improve factual consistency and diagnostic accuracy, recent studies have explored knowledge-enhanced report generation. Knowledge graph–based methods~\citep{liu2021auto} introduce structured medical entity relations to support clinically grounded reasoning, while retrieval-augmented generation (RAG) approaches incorporate retrieved reports or diagnosis-aware prompts into the generation process~\citep{ranjit2023retrieval,jin2024promptmrg}. Nevertheless, these methods are mainly developed for chest X-ray report generation and rarely consider mammography-specific diagnostic workflows and clinical priors. In mammography, visually similar cases may correspond to substantially different BI-RADS assessments, making generic similarity-based retrieval insufficient for capturing clinically critical diagnostic information.

In contrast, our method explicitly incorporates mammography-specific clinical priors into the report generation process to better model the structured reasoning workflow required in mammography reporting. This design enables more clinically consistent and diagnostically accurate report generation.

\subsection{Automated Report Parsing Tools}
Automated report parsing tools are essential for evaluating clinical correctness beyond surface-level text similarity metrics. Existing tools, such as CheXpert~\citep{irvin2019chexpert}, CheXbert~\citep{smit2020combining}, and RadGraph-based methods~\citep{NEURIPSDATASETSANDBENCHMARKS2021_c8ffe9a5,yan2023attributed}, are primarily designed for chest X-ray report evaluation by extracting predefined clinical entities and relations.

However, these tools are not directly applicable to mammography due to substantial differences in anatomical structures, BI-RADS assessment systems, and mammography-specific clinical terminology. Therefore, to facilitate reliable CE evaluation for mammography report generation, we developed MammoRGTool, a dedicated mammography report parsing tool that extracts structured clinical information from free-text mammography reports.

\section{Method}
In this section, we first describe the overall framework of MammoRG, followed by the proposed two-stage training strategy, namely cross-modal clinical knowledge integration and term-aware supervised fine-tuning. Finally, we introduce our report parsing tool, MammoRGTool.

\begin{figure}
\centering
\includegraphics[width=\textwidth]{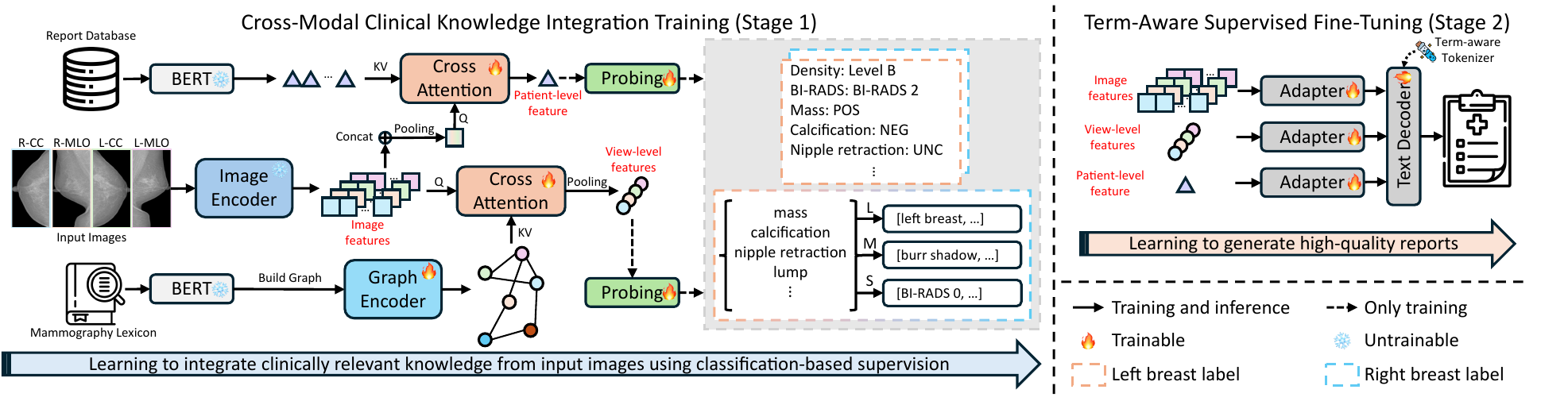}
\caption{\textbf{Overview of MammoRG.} This figure illustrates the two-stage training process and the inference workflow, where $L$, $M$, and $S$ in stage 1 represent \textit{Located\_at}, \textit{Modified\_by}, and \textit{Suggestive\_of}, respectively.} 
\label{fig:mammorg}
% \vspace{-1.5em}
\end{figure}

\subsection{Framework}

The overall architecture of our framework is illustrated in Fig. \ref{fig:mammorg}. 
Our method adopts an encoder--decoder architecture for mammography report generation. 
The input consists of four standard mammography views, namely 
$\{\text{R-CC}, \text{R-MLO}, \text{L-CC}, \text{L-MLO}\}$. These images are encoded by an image encoder ($E_{\text{img}}$) to extract image features:
\begin{equation}
\mathbf{v} = E_{\text{img}}\!\left(I^{\text{R-CC}}, I^{\text{R-MLO}}, I^{\text{L-CC}}, I^{\text{L-MLO}}\right).
\end{equation}

Clinically relevant prior knowledge from the report database and the mammography knowledge graph is incorporated at both the patient and view levels. 
The integrated clinical knowledge features are then fused with the image features and fed into a text decoder ($D_{\text{txt}}$) to generate the final mammography report:
\begin{equation}
\hat{Y} = D_{\text{txt}}(\mathbf{v}, \mathbf{k}),
\end{equation}
where $\mathbf{k}$ denotes the integrated clinical knowledge features.

\paragraph{Training Strategy.}
The training procedure consists of two stages. 
The first stage is the \emph{cross-modal clinical knowledge integration training stage}, which aims to learn clinically relevant diagnostic semantics from the report database and mammography knowledge graph under classification-based supervision. 
The second stage is the \emph{term-aware supervised fine-tuning stage}, where a term-aware tokenizer is introduced to model domain-specific clinical terms as atomic semantic units, enabling the model to generate high-quality reports based on the integrated multimodal features.

\subsection{Cross-Modal Clinical Knowledge Integration}
In medical image report generation, relying solely on image features is often insufficient to capture clinically critical information.
By incorporating auxiliary features that embed prior knowledge, models can generate reports that are more accurate and clinically meaningful.
In mammography report generation, radiologists typically analyze four standard views,
namely R-CC, R-MLO, L-CC, and L-MLO, and follow the BI-RADS guideline \citep{spak2017bi}
with prior clinical experience to produce diagnostic reports.

Inspired by this diagnostic workflow, MammoRG treats the mammography lexicon
as a structured representation of the BI-RADS guideline, and leverages all reports
in the training set as accumulated clinical experience.
By integrating clinically relevant prior knowledge (e.g., clinical experience and BI-RADS guideline) conditioned on the input images, MammoRG enriches visual representations with diagnostic semantic context, thereby improving report generation performance.

Rather than relying on image-similarity-based retrieval or generic image--text matching methods, our framework focuses on aligning visual features with clinically relevant diagnostic semantics. 
This design is motivated by two key observations. First, visually similar mammograms may correspond to substantially different diagnostic conclusions and BI-RADS assessments, making low-level visual similarity insufficient for capturing clinically critical information. Second, global image--text similarity often fails to distinguish clinically important semantics from non-critical contextual information, leading to suboptimal diagnostic alignment. 

In contrast, our framework integrates mammography-specific clinical priors into the report generation process through cross-modal semantic alignment. This design enables the model to incorporate clinically meaningful reasoning patterns rather than relying on superficial similarity, thereby improving both diagnostic accuracy and clinical consistency.

Specifically, we employ a pretrained BERT model \citep{devlin2019bert}
to encode all reports in the report database, yielding a set of report embeddings
$\mathbf{R} = \{ \mathbf{r}_1, \mathbf{r}_2, \dots, \mathbf{r}_N \}$,
where $N$ denotes the total number of reports.
Similarly, all mammography terms in the mammography lexicon are encoded into
term embeddings
$\mathbf{T} = \{ \mathbf{t}_1, \mathbf{t}_2, \dots, \mathbf{t}_M \}$,
where $M$ is the number of distinct mammography terms.

Based on the relation closure defined by MammoRGTool, we construct a mammography knowledge graph $\mathcal{G}$, in which each node corresponds to a term embedding
from $\mathbf{T}$ and edges represent predefined semantic relations, including $Located\_at$, $Modified\_by$, and $Suggestive\_of$.
The mammography knowledge graph is processed by a graph encoder composed of $L=4$
graph convolutional layers, resulting in graph-based knowledge representations  $\mathbf{G} = E_{\text{graph}}(\mathcal{G})$.

The visual input consists of four standard mammography views.
We define the set of mammography views as
\[
S_{\text{view}} = \{\text{R-CC}, \text{R-MLO}, \text{L-CC}, \text{L-MLO}\}.
\]
An image encoder $E_{\text{img}}(\cdot)$ is applied to each mammography view
to extract view-specific image features:
\begin{equation}
\mathbf{v}^{(i)} = E_{\text{img}}\!\left(I^{(i)}\right),
\quad i \in S_{\text{view}}.
\end{equation}

To construct a patient-level feature, we aggregate the features
from all mammography views and apply an average pooling operation:
\begin{equation}
\mathbf{v}_{\text{global}} =
\mathrm{AvgPool}\!\left(
\left\{
\mathbf{v}^{(i)} \mid i \in S_{\text{view}}
\right\}
\right).
\end{equation}

We employ cross-attention mechanisms to integrate clinically relevant semantic information at both the patient and view levels.
Specifically, the global image representation $\mathbf{v}_{\text{global}}$
is used as the query, while the report embeddings $\mathbf{R}$ serve as keys and values
to obtain patient-level clinical semantic representations:
\begin{equation}
\mathbf{f}_{\text{patient}} =
\mathrm{CrossAttn}(\mathbf{v}_{\text{global}}, \mathbf{R}).
\end{equation}
In contrast, each view-specific image feature $\mathbf{v}^{(i)}$ is employed
as a query to attend over the static mammography knowledge graph embeddings $\mathbf{G}$.
The resulting view-level features are then averaged across all views:
\begin{equation}
\mathbf{f}_{\text{view}} =
\mathrm{AvgPool}\!\left(
\left\{
\mathrm{CrossAttn}(\mathbf{v}^{(i)}, \mathbf{G})
\mid i \in S_{\text{view}}
\right\}
\right).
\end{equation}

Finally, the patient-level feature $\mathbf{f}_{\text{patient}}$ and the aggregated
view-level features $\mathbf{f}_{\text{view}}$ are separately fed into linear probing layers.
During training, structured category labels are extracted from ground-truth reports
using MammoRGTool as supervision signals, encouraging the model to learn clinically relevant diagnostic semantics conditioned on the input images. 

We optimize the model using standard classification objectives. Specifically, the cross-entropy loss $\mathcal{L}_{\text{CE}}$ is adopted for BI-RADS and breast composition classification, while the binary cross-entropy loss $\mathcal{L}_{\text{BCE}}$ is used for multi-label prediction of findings and relations. The overall objective for this stage is defined as:
\begin{equation}
\mathcal{L}_{\text{stage1}} = \mathcal{L}_{\text{CE}} + \mathcal{L}_{\text{BCE}}.
\end{equation}

\subsection{Term-Aware Supervised Fine-tuning}

Current models utilizing vision-language models for medical report generation, such as LLaVA-Rad \citep{zambrano2025clinically}, typically undergo a two-stage supervised fine-tuning process consisting of alignment and fine-tuning. The alignment stage focuses on aligning visual and textual representations, while the fine-tuning stage enables the language model to perform complex reasoning, engage in dialogue, and follow instructions based on the observed visual information. However, we find that existing tokenizers for language models, such as WordPiece used in BERT \citep{devlin2019bert} and Byte Pair Encoding (BPE) in GPT \citep{radford2019language}, are trained on general web corpora to represent text efficiently with a limited vocabulary. As a result, domain-specific terms are often fragmented into multiple subwords, which introduces two drawbacks. First, the model must infer the semantics of a clinical term from its constituent pieces, increasing sample complexity and potentially introducing semantic ambiguity for clinically critical concepts. Second, subword fragmentation inflates sequence length, increasing computation and memory usage and potentially degrading long-range modeling by consuming context capacity.

In our method, we propose a term-aware tokenizer that treats each term from the mammography lexicon as an indivisible token. The language model is explicitly trained to learn these tokens as domain-specific semantic units, thereby preserving the atomic semantic integrity of clinical concepts and facilitating more direct alignment between visual features and structured medical knowledge. This eliminates the need for the model to infer semantics from subword compositions and reduces semantic fragmentation. Consequently, the model achieves more consistent predictions on clinically relevant attributes, while also benefiting from shorter input sequences and improved computational efficiency.

Although the newly introduced tokens are not associated with pretrained embeddings from the original language model, they are initialized in a data-driven manner and further optimized during supervised fine-tuning. Specifically, we initialize each new token embedding using the mean of existing embeddings with small random perturbations, providing a stable starting point that is consistent with the original embedding space. During training, these embeddings are updated through standard language modeling objectives based on report generation, where domain-specific terms frequently appear in context. This allows the model to gradually learn their semantic representations from contextual usage, rather than relying on explicit pretraining. Moreover, the use of parameter-efficient tuning (LoRA \citep{hu2022lora}) helps preserve the pretrained backbone while allowing the newly introduced embeddings to be adapted without disrupting existing knowledge. As a result, the proposed tokenizer integrates smoothly with the pretrained model and supports effective learning of domain-specific terminology.

We optimize the model using the standard autoregressive language modeling objective $\mathcal{L}_{\text{stage2}}$ for report generation.

\subsection{MammoRGTool}

\begin{figure}
\centering
\includegraphics[width=\textwidth]{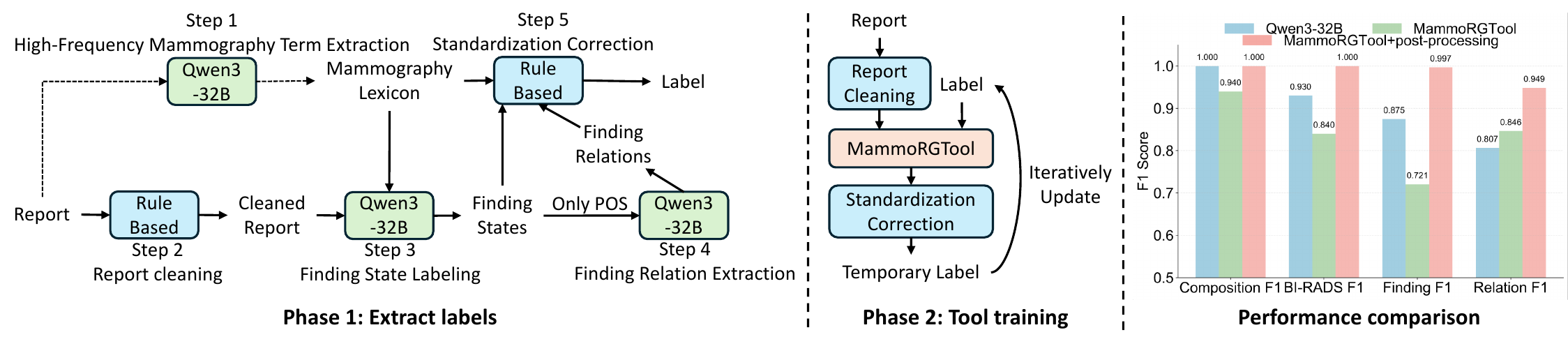}
\caption{\textbf{Overview of MammoRGTool.} This figure illustrates the two phases of MammoRGTool development and compares the performance of Qwen3-32B, MammoRGTool, and MammoRGTool with post-processing on 50 manually annotated samples.}
\label{fig:mammorgtool}
% \vspace{-1.5em}
\end{figure}

To develop \textit{MammoRGTool}, we first leverage Qwen3-32B \citep{yang2025qwen3} to extract BI-RADS categories, tissue composition types, abnormal findings, along with the associated anatomical locations, descriptive terms, and diagnostic suggestions for bilateral breasts across standard mammographic views (e.g., CC and MLO) from all collected mammography reports. These extracted results are then used as supervisory labels to iteratively train MammoRGTool, as illustrated in Fig.~\ref{fig:mammorgtool}.

The label extraction pipeline begins with applying Qwen3-32B to extract candidate entities from reports, after which we compute entity frequencies and construct a high-frequency, non-redundant mammography lexicon. This lexicon comprises 9 BI-RADS categories, 4 tissue composition types, 10 common abnormal findings, 18 standard anatomical locations, 28 typical descriptors, and 10 diagnostic suggestions (see Appendix). After removing anomalies and extraneous characters, we use Qwen3-32B to identify the presence states of the 10 abnormal findings, including Positive (POS), Negative (NEG), Uncertain (UNC), or Blank (BLA). For positively identified findings, we further extract three types of relations: location (\textit{Located\_at}), modification (\textit{Modified\_by}), and suggestion (\textit{Suggestive\_of}). Finally, all extracted entities and relations are constrained to the predefined mammography lexicon to form the final labels.

In the tool training phase, Qwen3-32B is replaced by MammoRGTool, while the overall pipeline remains unchanged. MammoRGTool adopts OneRel \citep{shang2022onerel} for relation extraction and employs six classification heads to independently predict BI-RADS categories, tissue composition types, and abnormal finding presence for bilateral breasts. Training proceeds in an iterative self-refinement manner, where every five epochs, newly predicted labels are used as updated supervision to progressively mitigate noise introduced by Qwen3-32B.

Evaluation on 50 manually annotated reports (as shown in the right of Fig.~\ref{fig:mammorgtool}) shows that, while MammoRGTool slightly underperforms Qwen3-32B in extracting composition, BI-RADS categories, and finding presence, it achieves superior performance in relation extraction. In addition, a rule-based post-processing stage is applied to further improve the overall extraction accuracy.

\section{Experiments}
\subsection{Datasets}

Due to the limited availability of public mammography report datasets, we curate a multi-center dataset comprising paired four-view mammography images and reports from three medical centers. Data from Yunnan Cancer Hospital is used for internal training, validation, and testing, with 11,776, 1,733, and 3,441 pairs, respectively. Data from the remaining two centers (Guangzhou First People’s Hospital and The Third Affiliated Hospital, Sun Yat-Sen University) serve as two external validation sets, containing 1,239 and 1,153 pairs. In addition, we include the public VinDr-Mammo dataset \citep{nguyen2023vindr} as an external validation set. VinDr-Mammo contains 5,000 cases with complete four-view mammography images and categorical annotations, including BI-RADS categories, breast composition, and abnormal findings. As reports are not provided, we apply MammoRGTool to extract categorical predictions from generated reports and compute CE metrics directly against the annotated labels.

\subsection{Evaluation Metrics}
\label{sec:eval_metrics}
We evaluate model performance with both conventional natural language generation (NLG) metrics and the proposed CE metrics. The NLG metrics include BLEU-1 \citep{papineni2002bleu} and ROUGE-L \citep{lin2004rouge}, which mainly measure lexical overlap between generated and reference reports. The CE metrics are four macro-F1 scores computed by MammoRGTool, covering breast composition type, BI-RADS category, presence of abnormal findings, and relations among findings, and they quantify agreement on clinically relevant structured attributes.
\textbf{Breast Composition Type and BI-RADS Category.}
Breast composition type and BI-RADS category are evaluated as multi-class classification tasks. Predictions are collected from both the left and right breasts across the entire dataset. Only cases where the ground-truth label is not \textit{BLA} are considered evaluable. Predictions of \textit{BLA} are treated as missing and accounted for through a global completeness penalty.

Let $\mathcal{C}$ denote the class set (breast composition or BI-RADS). For each class $c \in \mathcal{C}$, precision and recall are defined as:
\begin{equation}
P_c = \frac{\mathrm{TP}_c}{\mathrm{TP}_c + \mathrm{FP}_c},
\qquad
R_c = \frac{\mathrm{TP}_c}{\mathrm{TP}_c + \mathrm{FN}_c}.
\end{equation}
The final completeness-aware macro-F1 score is computed as:
\begin{equation}
\mathrm{F1}
=
\left(
\frac{1}{|\mathcal{C}|}
\sum_{c \in \mathcal{C}}
\frac{2 P_c R_c}{P_c + R_c}
\right)
\cdot
\frac{N^{\mathrm{actual}}}{N^{\mathrm{should}}}.
\end{equation}
Here, $N^{\mathrm{should}}$ denotes the total number of evaluable ground-truth labels in the dataset, and $N^{\mathrm{actual}}$ denotes the number of corresponding non-\textit{BLA} predictions produced by the model. This formulation is applied identically to bilateral breast composition type and BI-RADS category.

\textbf{Presence of Abnormal Findings.}
The presence of abnormal findings is evaluated as a binary classification task, where only the positive class (\textit{POS}) is explicitly assessed. Predictions are aggregated across all predefined entities and bilateral breasts. \textit{NEG}, \textit{BLA} and \textit{UNC} are treated as the negative class. Precision and recall for the positive class are defined as:
\begin{equation}
P^{+} = \frac{\mathrm{TP}^{+}}{\mathrm{TP}^{+} + \mathrm{FP}^{+}},
\qquad
R^{+} = \frac{\mathrm{TP}^{+}}{\mathrm{TP}^{+} + \mathrm{FN}^{+}}.
\end{equation}
The F1-score for abnormal findings is then computed as:
\begin{equation}
\mathrm{F1}^{\mathrm{entities}}
=
\frac{2 P^{+} R^{+}}{P^{+} + R^{+}}.
\end{equation}

\textbf{Relations of Findings.}
Relations of findings are evaluated by comparing the predicted and reference sets of relational triples aggregated across the test dataset. Let $\mathcal{T}^{\mathrm{pred}}$ and $\mathcal{T}^{\mathrm{ref}}$ denote the sets of predicted and ground-truth triples, respectively. Precision and recall for relation extraction are defined as:
\begin{equation}
P^{\mathrm{rel}} = 
\frac{
\left| \mathcal{T}^{\mathrm{pred}} \cap \mathcal{T}^{\mathrm{ref}} \right|
}{
\left| \mathcal{T}^{\mathrm{pred}} \right|
},
\qquad
R^{\mathrm{rel}} =
\frac{
\left| \mathcal{T}^{\mathrm{pred}} \cap \mathcal{T}^{\mathrm{ref}} \right|
}{
\left| \mathcal{T}^{\mathrm{ref}} \right|
}.
\end{equation}
The relation-level F1-score is computed as:
\begin{equation}
\mathrm{F1}^{\mathrm{relations}}
=
\frac{2 P^{\mathrm{rel}} R^{\mathrm{rel}}}
{P^{\mathrm{rel}} + R^{\mathrm{rel}}}.
\end{equation}

\subsection{Implementation Details}

We adopt VersaMammo \citep{huang2025versatile}, a mammography-specific foundation model, as the image encoder, and LLaVA-Mammo \citep{zhu2025benchmark} as the text decoder. Since all reports are written in Chinese, a Chinese-adapted BERT \citep{devlin2019bert} is employed to encode the report database and mammography lexicon.

During the cross-modal clinical knowledge integration training stage, all the current parameters, including the graph encoder, cross-attention, and probing layers, are trained. During the term-aware supervised fine-tuning stage, the model focuses on adapting the language generation capability to mammography-specific terminology and reporting conventions. In the alignment phase, only the adapters, embedding layer, and language modeling head are updated to efficiently align the pretrained language model with the expanded mammography lexicon. In the subsequent fine-tuning phase, LoRA \citep{hu2022lora} ($r=64$, $\alpha=16$) is further applied to the text decoder to improve report generation quality while maintaining parameter-efficient adaptation.

Training uses 100, 1, and 3 epochs for the cross-modal clinical knowledge integration, alignment, and fine-tuning stages, respectively. A longer training schedule is adopted for the cross-modal clinical knowledge integration stage to ensure sufficient convergence of clinical semantic alignment learning, while the alignment and fine-tuning stages follow the training configuration of LLaVA-Rad \citep{zambrano2025clinically}. The corresponding batch sizes are 64, 16, and 4, respectively, which are selected to maximize GPU utilization under different computational demands across training stages. Initial learning rates are set to 1e-4, 1e-3, and 1e-4 under a cosine learning rate schedule. All images are resized to 512$\times$512 to preserve fine-grained mammographic details that are critical for accurate diagnosis. The model is implemented in the LLaVA-Rad environment and trained on a single NVIDIA A100 GPU for approximately 40 hours.

\subsection{Results}

\subsubsection{Quantitative comparison} 

We compare our model with other state-of-the-art (SOTA) models, including three large vision-language models (LVLMs): Qwen3-VL-8B \citep{Qwen3-VL}, InternVL3.5-8B \citep{wang2025internvl3}, and Lingshu-7B \citep{xu2025lingshu}. These models were tested in a one-shot setting, with a single example provided in the prompt. Additionally, we retrain LLaVA-Rad \citep{zambrano2025clinically}, Mammo2Text \citep{yalunin2021generating}, R2Gen \citep{chen-emnlp-2020-r2gen}, M2KT \citep{yang2023radiology}, and PromptMRG \citep{jin2024promptmrg} on the training set of MammoRG, using the code released by the respective authors. For fair comparison, the number of training epochs for all models was set to 4, consistent with our baseline (LLaVA-Rad), and the image encoder in each model is replaced with VersaMammo.

\begin{table*}[t]
\centering
\scriptsize
\caption{Comparison with SOTA methods on the internal test dataset, two external test datasets, and VinDr-Mammo. All evaluation metrics are reported in percentage (\%) form. `One-shot' refers to the vision-language models that use a prompt method with a given report example for reasoning. `Trained' indicates the model after training on our internal training dataset. The best performing model for
each metric is \textbf{bolded} and the second-best performing model is \underline{underlined}. The standard deviation (std) is estimated via 1000 bootstrap resampling runs.}
\label{tab:quantitative_results}
\setlength{\tabcolsep}{5pt}
\resizebox{\textwidth}{!}{
\begin{tabular}{llllll|llllll}
\toprule
\multirow{2}{*}{Dataset} & \multirow{2}{*}{Metrics Type} & \multirow{2}{*}{Metrics} & \multicolumn{3}{c|}{One-shot} & \multicolumn{6}{c}{Trained} \\
& & & Qwen3-VL-8B & InternVL3.5-8B & Lingshu-7B & LLaVA-Rad & Mammo2Text & R2Gen & M2KT & PromptMRG & MammoRG (Ours) \\
\midrule

\multirow{6}{*}{Internal} 
& \multirow{2}{*}{NLG Metrics} 
& BLEU-1   & 39.56$\pm$0.22 & 36.66$\pm$0.24 & 26.37$\pm$0.40 & \underline{52.37$\pm$0.34} & 47.90$\pm$0.35 & 52.28$\pm$0.30 & 44.81$\pm$0.26 & 51.48$\pm$0.33 & \textbf{53.78$\pm$0.34} \\
& & ROUGE-L & 38.08$\pm$0.20 & 34.24$\pm$0.22 & 27.60$\pm$0.29 & \underline{61.60$\pm$0.24} & 58.97$\pm$0.30 & 58.84$\pm$0.29 & 52.18$\pm$0.24 & 60.52$\pm$0.26 & \textbf{62.70$\pm$0.26} \\
% \cline{2-12}
& \multirow{4}{*}{CE Metrics} 
& Composition F1 & 21.28$\pm$0.41 & 15.79$\pm$0.50 & 7.55$\pm$0.27 & 41.61$\pm$2.20 & 46.68$\pm$2.17 & 41.29$\pm$2.52 & \underline{47.02$\pm$2.05} & 37.15$\pm$1.18 & \textbf{47.79$\pm$2.29} \\
& & BI-RADS F1   & 4.65$\pm$0.20  & 1.39$\pm$0.13  & 0.02$\pm$0.01 & \underline{23.94$\pm$0.50} & 19.95$\pm$1.23 & 13.07$\pm$0.55 & 14.30$\pm$0.52 & 21.20$\pm$1.02 & \textbf{26.67$\pm$0.63} \\
& & Finding F1   & 23.11$\pm$0.43 & 8.76$\pm$0.33  & 7.68$\pm$0.41 & \underline{72.13$\pm$0.44} & 65.81$\pm$0.56 & 68.05$\pm$0.51 & 64.11$\pm$0.47 & 71.58$\pm$0.52 & \textbf{75.09$\pm$0.38} \\
& & Relation F1  & 4.75$\pm$0.45  & 1.65$\pm$0.47  & 1.00$\pm$0.44 & 37.23$\pm$0.26 & 35.75$\pm$0.29 & 34.10$\pm$0.26 & 28.46$\pm$0.17 & \underline{40.14$\pm$0.29} & \textbf{40.85$\pm$0.30} \\
\midrule

\multirow{6}{*}{External 1} 
& \multirow{2}{*}{NLG Metrics} 
& BLEU-1   & 35.69$\pm$0.34 & 30.87$\pm$0.28 & 21.98$\pm$0.58 & 36.11$\pm$0.37 & \underline{37.38$\pm$0.29} & 25.81$\pm$0.36 & 27.29$\pm$0.33 & 36.29$\pm$0.55 & \textbf{38.78$\pm$0.48} \\
& & ROUGE-L & 34.75$\pm$0.37 & 30.50$\pm$0.34 & 23.43$\pm$0.29 & \underline{48.46$\pm$0.31} & 40.68$\pm$0.39 & 35.27$\pm$0.36 & 36.90$\pm$0.31 & 42.85$\pm$0.30 & \textbf{49.17$\pm$0.33} \\
% \cline{2-12}
& \multirow{4}{*}{CE Metrics} 
& Composition F1 & 16.87$\pm$0.58 & 15.00$\pm$0.96 & 7.15$\pm$0.40 & 35.85$\pm$1.97 & \underline{39.03$\pm$2.30} & 31.56$\pm$1.69 & 36.35$\pm$1.89 & \textbf{39.62$\pm$2.01} & 37.93$\pm$2.02 \\
& & BI-RADS F1   & 3.78$\pm$0.33  & 1.42$\pm$0.23  & 0.06$\pm$0.04 & \underline{12.07$\pm$0.57} & 7.51$\pm$0.49  & 6.44$\pm$0.50  & 7.17$\pm$0.52  & 7.85$\pm$0.50  & \textbf{14.11$\pm$0.64} \\
& & Finding F1   & 29.73$\pm$0.83 & 11.26$\pm$0.69 & 9.38$\pm$0.68 & 36.75$\pm$0.79 & \underline{40.30$\pm$0.88} & 30.49$\pm$0.75 & 36.17$\pm$0.83 & \textbf{41.42$\pm$0.87} & 37.34$\pm$0.81 \\
& & Relation F1  & 6.12$\pm$0.71  & 2.41$\pm$0.89  & 1.70$\pm$0.95 & \underline{8.33$\pm$0.31}  & 7.89$\pm$0.28  & 5.19$\pm$0.29  & 7.05$\pm$0.25  & 7.27$\pm$0.22  & \textbf{8.46$\pm$0.37} \\
\midrule

\multirow{6}{*}{External 2} 
& \multirow{2}{*}{NLG Metrics} 
& BLEU-1   & 39.68$\pm$0.45 & 35.85$\pm$0.45 & 20.67$\pm$0.78 & \underline{42.52$\pm$0.43} & 32.90$\pm$0.68 & 42.33$\pm$0.54 & \textbf{43.42$\pm$0.43} & 42.22$\pm$0.47 & 41.96$\pm$0.49 \\
& & ROUGE-L & 44.94$\pm$0.40 & 41.33$\pm$0.43 & 31.59$\pm$0.44 & 48.43$\pm$0.37 & 35.89$\pm$0.52 & \textbf{54.60$\pm$0.64} & 42.87$\pm$0.26 & 47.99$\pm$0.35 & \underline{50.49$\pm$0.38} \\
% \cline{2-12}
& \multirow{4}{*}{CE Metrics} 
& Composition F1 & 21.64$\pm$0.77 & 17.66$\pm$0.89 & 7.09$\pm$0.41 & 44.43$\pm$2.66 & 42.45$\pm$2.60 & \textbf{53.26$\pm$3.26} & 47.75$\pm$2.84 & 45.68$\pm$1.82 & \underline{50.88$\pm$1.24} \\
& & BI-RADS F1   & 4.72$\pm$0.62  & 0.92$\pm$0.17  & 0.00$\pm$0.00 & \underline{13.50$\pm$0.87} & 9.93$\pm$0.60  & 10.95$\pm$0.89 & 9.64$\pm$0.79  & 10.84$\pm$0.72 & \textbf{15.40$\pm$1.00} \\
& & Finding F1   & 31.31$\pm$1.08 & 10.71$\pm$1.06 & 3.48$\pm$0.66 & 31.17$\pm$0.99 & \textbf{34.94$\pm$1.15} & 22.16$\pm$0.88 & 23.68$\pm$0.98 & \underline{33.92$\pm$1.12} & 31.40$\pm$0.94 \\
& & Relation F1  & 6.96$\pm$1.65  & 2.34$\pm$2.45  & 0.50$\pm$2.21 & 6.82$\pm$0.43  & 5.53$\pm$0.54  & 5.08$\pm$0.72  & 5.73$\pm$0.35  & \underline{7.00$\pm$0.49}  & \textbf{7.61$\pm$0.55} \\
\midrule

\multirow{3}{*}{VinDr-Mammo} 
& \multirow{3}{*}{CE Metrics} 
& Composition F1 & 21.70$\pm$0.35 & 20.81$\pm$0.48 & 5.21$\pm$0.18 & 50.71$\pm$1.43 & 46.73$\pm$1.00 & 37.91$\pm$1.75 & 40.24$\pm$1.42 & \underline{54.17$\pm$1.33} & \textbf{54.76$\pm$1.57} \\
& & BI-RADS F1   & 8.76$\pm$0.35  & 4.58$\pm$0.32  & 0.00$\pm$0.01 & 26.36$\pm$0.60 & 25.68$\pm$0.67 & \underline{27.80$\pm$0.66} & 18.41$\pm$0.55 & 26.11$\pm$0.61 & \textbf{31.07$\pm$0.79} \\
& & Finding F1   & 5.02$\pm$0.34  & 4.16$\pm$0.47  & 2.10$\pm$0.47 & 8.05$\pm$0.62  & 6.75$\pm$0.47  & 8.35$\pm$0.52  & 6.74$\pm$0.33  & \textbf{9.18$\pm$0.58}  & \underline{8.71$\pm$0.61} \\
\midrule

\multirow{6}{*}{Average}
& \multirow{2}{*}{NLG Metrics}
& BLEU-1
& 38.31$\pm$0.34 & 34.46$\pm$0.32 & 23.01$\pm$0.59 & \underline{43.67$\pm$0.38} & 39.39$\pm$0.44 & 40.14$\pm$0.40 & 38.51$\pm$0.34 & 43.33$\pm$0.45 & \textbf{44.84$\pm$0.44} \\

& & ROUGE-L
& 39.26$\pm$0.32 & 35.36$\pm$0.33 & 27.54$\pm$0.34 & \underline{52.83$\pm$0.31} & 45.18$\pm$0.40 & 49.57$\pm$0.43 & 43.98$\pm$0.27 & 50.45$\pm$0.30 & \textbf{54.12$\pm$0.32} \\

% \cline{2-12}

& \multirow{4}{*}{CE Metrics}

& Composition F1
& 20.37$\pm$0.53 & 17.32$\pm$0.71 & 6.75$\pm$0.32 & 43.15$\pm$2.06 & 43.72$\pm$2.02 & 41.00$\pm$2.30 & 42.84$\pm$2.05 & \underline{44.16$\pm$1.58} & \textbf{47.84$\pm$1.78} \\

&& BI-RADS F1
& 5.48$\pm$0.38 & 2.08$\pm$0.21 & 0.02$\pm$0.02 & \underline{18.97$\pm$0.64} & 15.77$\pm$0.75 & 14.57$\pm$0.65 & 12.38$\pm$0.59 & 16.50$\pm$0.71 & \textbf{21.81$\pm$0.77} \\

&& Finding F1
& 22.29$\pm$0.67 & 8.72$\pm$0.64 & 5.66$\pm$0.55 & 37.03$\pm$0.71 & 36.95$\pm$0.76 & 32.26$\pm$0.67 & 32.68$\pm$0.65 & \textbf{39.03$\pm$0.77} & \underline{38.14$\pm$0.68} \\

&& Relation F1
& 5.94$\pm$0.94 & 2.13$\pm$1.27 & 1.07$\pm$1.20 & 17.46$\pm$0.33 & 16.39$\pm$0.37 & 14.79$\pm$0.42 & 13.75$\pm$0.26 & \underline{18.14$\pm$0.33} & \textbf{18.97$\pm$0.41} \\

\bottomrule
\end{tabular}}
\end{table*}

Tab. \ref{tab:quantitative_results} reports the performance of these methods on the four datasets. Overall, one-shot LVLMs can achieve moderate NLG scores, yet their CE performance, especially for BI-RADS, remains extremely low, suggesting that one-shot prompting is insufficient to produce clinically actionable structure. By contrast, retraining prior report-generation methods under a unified protocol substantially improves both NLG and CE scores. 
MammoRG consistently delivers the strongest overall performance. On the internal test set, it achieves the best NLG scores (i.e., BLEU-1: 53.78\%; ROUGE-L: 62.70\%) and attains the highest composition F1 (47.79\%), BI-RADS F1 (26.67\%), Finding F1 (75.09\%), and Relation F1 (40.85\%), outperforming the second-best model (LLaVA-Rad) by 1.41\%, 1.10\%, 6.18\%, 2.73\%, 2.96\%, and 3.62\%, respectively. On External 1, MammoRG attains the highest BLEU-1 (38.78\%), ROUGE-L (49.17\%), BI-RADS F1 (14.11\%), and Relation F1 (8.46\%), while its Composition F1 and Finding F1 are lower than PromptMRG, which may reflect domain shift (e.g., differences in reporting style and label distribution) that more strongly impacts breast composition and finding-presence criteria. On External 2, MammoRG still maintains stable generalization performance compared to other models (e.g., PromptMRG). It achieves the best results on BI-RADS F1 (15.40\%) and Relation F1 (7.61\%). Its Composition F1 (50.88\%) is 2.38\% lower than the best-performing model, R2Gen (53.26\%). In contrast, LLaVA-Rad and PromptMRG, which perform strongly on the internal test set and External 1, achieve only 44.43\% and 45.68\%, respectively. On the public external dataset, VinDr-Mammo, where the NLG metrics are not available, MammoRG achieves the best Composition F1 (54.76\%) and BI-RADS F1 (31.07\%), outperforming the second-best model (PromptMRG) by 0.59\% and 4.96\%, respectively.

Across all datasets, MammoRG achieves the best average performance on five out of six evaluation metrics, including BLEU-1 (44.84\%), ROUGE-L (54.12\%), Composition F1 (47.84\%), BI-RADS F1 (21.81\%), and Relation F1 (18.97\%). For Finding F1, MammoRG achieves the second-best average score (38.14\%), only slightly lower than PromptMRG (39.03\%). These results further demonstrate the strong robustness and cross-domain generalization capability of MammoRG across diverse clinical distributions and reporting styles.

Notably, MammoRG shows the most consistent gains on CE metrics directly tied to downstream decision-making (e.g., BI-RADS). These results indicate that our method exhibits superior comprehensive analytical and reasoning capabilities compared to existing models, enabling more accurate benign–malignant diagnoses for patients.

\begin{figure}
\centering
\includegraphics[width=\textwidth]{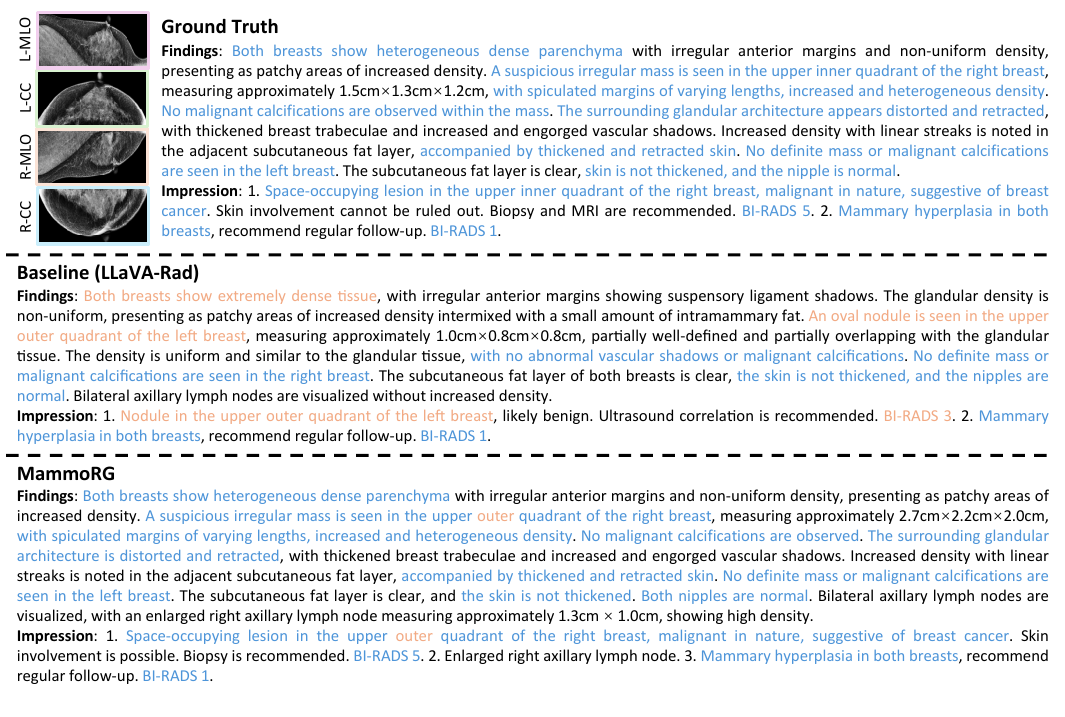}
\caption{Qualitative examples of the generated report of the baseline and the proposed method. Blue font indicates consistent content with the ground-truth, while red font indicates incorrect content.}
\label{fig:qualitative}
% \vspace{-1.5em}
\end{figure}

\subsubsection{Qualitative comparison}

We present a qualitative comparison to illustrate the superiority of MammoRG over the baseline model (LLaVA-Rad), as shown in Fig. \ref{fig:qualitative}. Blue text denotes predictions consistent with the ground truth, while red text indicates errors. MammoRG accurately captures most clinically critical information in the ground truth report, including breast composition, abnormal findings such as mass morphology and architectural distortion, and bilateral BI-RADS assessments, with only a minor deviation in mass location. In contrast, although the baseline model employs professional terminology, it fails to correctly predict key clinical attributes, misclassifying breast composition and the severity and location of the mass, which leads to an incorrect BI-RADS assessment. 

\subsubsection{Ablation study}

In this section, we conduct ablation studies from the following three perspectives:

(1) \textbf{Feature combination.}

We investigate the impact of different combinations of the three feature types—\emph{image features}, \emph{patient-level features}, and \emph{view-level features}—on the quality of generated reports. As shown in Tab. \ref{tab:feature_combination}, several key observations can be drawn: a) There are significant differences in the representational capacity of individual features. Image features achieve the best performance among all single-feature settings, outperforming the other two in both NLG metrics (e.g., BLEU-1: 53.30\%; ROUGE-L: 62.46\%) and Finding F1 (74.57\%), indicating that they contain the richest semantic and visual information. In contrast, patient-level and view-level features perform substantially worse when used alone, particularly in BI-RADS F1 and Relation F1, suggesting that relying solely on low-dimensional global representations is insufficient for generating comprehensive clinical descriptions. b) Features at different granularities exhibit strong complementarity. Combining either patient-level or view-level features with image features consistently improves performance. For instance, “image + patient-level” increases Composition F1 to 45.61\% and achieves the best Finding F1 (75.33\%), while “image + view-level” attains the best ROUGE-L (62.88\%), along with competitive BI-RADS F1 (25.96\%) and Relation F1 (40.49\%). This indicates that patient-level features enhance overall semantic consistency, whereas view-level features contribute more to fine-grained relational modeling. c) The combination of patient-level and view-level features demonstrates high efficiency with minimal token cost. Using only 5 tokens, this setting achieves competitive performance in both NLG (e.g., BLEU-1: 53.83\%) and CE metrics (e.g., Composition F1: 44.67\%), even surpassing some configurations that include image features. This highlights its high information density and computational efficiency. d) Integrating all three feature types yields the best overall performance. As indicated by the gray row, “image + patient-level + view-level” achieves the best Composition F1 (47.79\%), BI-RADS F1 (26.67\%), and Relation F1 (40.85\%), along with competitive Finding F1 (75.09\%), demonstrating that multi-granularity feature fusion significantly improves both generation quality and clinical consistency.

Overall, these results show that the three feature types provide complementary information at different levels, and that well-designed feature combinations can effectively balance performance and efficiency.

(2) \textbf{Component analysis.} 

To analyze the contribution of individual components, we conduct a stepwise ablation study starting from a baseline model (LLaVA-Rad) and progressively incorporating two components: the term-aware tokenizer and cross-modal clinical knowledge integration. The results are summarized in Tab. \ref{tab:component_ablation}, from which we draw the following observations: a) The term-aware tokenizer, which treats medical terms as atomic semantic units, consistently improves model performance. Compared with the baseline, it boosts NLG metrics (BLEU-1: +0.93\%, ROUGE-L: +0.86\%) and significantly enhances clinically relevant CE metrics, particularly Finding F1 (+2.44\%) and Relation F1 (+2.35\%). These gains indicate that reducing semantic fragmentation of medical terminology enables the model to better capture structured clinical information, rather than merely improving surface-level fluency. b) Cross-modal clinical knowledge integration further improves overall performance when combined with the term-aware tokenizer. As shown in the gray row, this combination achieves the best results on all the metrics, including BLEU-1 (53.78\%), ROUGE-L (62.70\%), Composition F1 (47.79\%), BI-RADS F1 (26.67\%), Finding F1 (75.09\%), and Relation F1 (40.85\%). These results demonstrate that cross-modal clinical knowledge integration effectively enriches the model with relevant clinical knowledge, leading to more accurate and clinically consistent report generation.

\begin{table*}[t]
    \centering
    \tiny
    \caption{Ablation study of feature combinations on the internal test set. The standard deviation (std) is estimated via 1000 bootstrap resampling runs.}
    \vspace{-0.5em}
    \setlength{\tabcolsep}{.5mm}
    \resizebox{\textwidth}{!}{
    \begin{tabular}{clccccccc}
    \toprule
    \multirow{2}{*}{\shortstack{Number of\\features}} & 
    \multirow{2}{*}{Feature combination} & 
    \multirow{2}{*}{\shortstack{Number of\\tokens}} & 
    \multicolumn{2}{c}{\shortstack{NLG Metrics}} & 
    \multicolumn{4}{c}{\shortstack{CE Metrics}} \\
    & & & BLEU-1 & ROUGE-L & Composition F1 & BI-RADS F1 & Finding F1 & Relation F1 \\
    \midrule

    \multirowcell{3}{1} 
    & Image features & 1024 & 53.30$\pm$0.32 & 62.46$\pm$0.25 & 41.22$\pm$1.90 & 24.29$\pm$0.53 & 74.57$\pm$0.37 & 39.58$\pm$0.26 \\
    & Patient-level feature & 1 & 45.62$\pm$0.33 & 47.04$\pm$0.28 & 41.91$\pm$0.83 & 12.20$\pm$0.67 & 56.19$\pm$0.45 & 22.12$\pm$0.17 \\
    & View-level features & 4 & 51.39$\pm$0.35 & 58.58$\pm$0.27 & 37.61$\pm$2.89 & 16.01$\pm$0.88 & 65.57$\pm$0.50 & 31.03$\pm$0.23 \\

    \midrule
    \multirowcell{3}{2} 
    & Image features + Patient-level feature & 1025 & 53.20$\pm$0.36 & 62.53$\pm$0.26 & 45.61$\pm$2.03 & 25.17$\pm$0.56 & \textbf{75.33$\pm$0.39} & 40.13$\pm$0.26 \\
    & Image features + View-level features & 1028 & 53.53$\pm$0.95 & \textbf{62.88$\pm$0.26} & 43.62$\pm$1.97 & 25.96$\pm$0.58 & 74.98$\pm$0.40 & 40.49$\pm$0.27 \\
    & Patient-level feature + View-level features & 5 & \textbf{53.83$\pm$0.29} & 61.82$\pm$0.28 & 44.67$\pm$2.16 & 24.22$\pm$0.52 & 74.19$\pm$0.39 & 38.42$\pm$0.26 \\

    \midrule
    \multirowcell{1}{3} 
    & \cellcolor{gray!10}Image features + Patient-level feature + View-level features
    & 1029 & 53.78$\pm$0.34 & 62.70$\pm$0.26 & \textbf{47.79$\pm$2.29} & \textbf{26.67$\pm$0.63} & 75.09$\pm$0.38 & \textbf{40.85$\pm$0.30} \\

    \bottomrule
    \end{tabular}}
    \label{tab:feature_combination}
\end{table*}

\begin{table*}[t]
    \centering
    \tiny
    \caption{Ablation study of different components on the internal test dataset. The standard deviation (std) is estimated via 1000 bootstrap resampling runs.}
    \vspace{-0.5em}
    \resizebox{\textwidth}{!}{
    \begin{tabular}{cccccccc}
    \toprule
    \multirow{2}{*}{\shortstack{Term-aware\\tokenizer}} & 
    \multirow{2}{*}{\shortstack{Cross-modal\\clinical knowledge integration}} & 
    \multicolumn{2}{c}{\shortstack{NLG Metrics}} & 
    \multicolumn{4}{c}{\shortstack{CE Metrics}} 
    \\
    & &  BLEU-1 & ROUGE-L & Composition F1 & BI-RADS F1 & Finding F1 & Relation F1 \\
    \midrule
    \XSolidBrush & \XSolidBrush &
    52.37$\pm$0.34 & 61.60$\pm$0.24 & 41.61$\pm$2.20 & 23.94$\pm$0.50 & 72.13$\pm$0.44 & 37.23$\pm$0.26  \\
    \midrule
    \CheckmarkBold & \XSolidBrush  &
    53.30$\pm$0.32 & 62.46$\pm$0.25 & 41.22$\pm$1.90 & 24.29$\pm$0.53 & 74.57$\pm$0.37 & 39.58$\pm$0.26  \\
    \midrule
    \cellcolor{gray!10}\CheckmarkBold & \cellcolor{gray!10}\CheckmarkBold &
    \textbf{53.78$\pm$0.34} & \textbf{62.70$\pm$0.26} & \textbf{47.79$\pm$2.29} & \textbf{26.67$\pm$0.63} & \textbf{75.09$\pm$0.38} & \textbf{40.85$\pm$0.30}  \\
    \bottomrule
    \end{tabular}}
    \label{tab:component_ablation}
\end{table*}

\begin{table*}[t]
    \centering
    \tiny
    \caption{Comparison between the proposed cross-modal clinical knowledge integration framework and different retrieval-based strategies on the internal test dataset. The standard deviation (std) is estimated via 1000 bootstrap resampling runs.}
    \vspace{-0.5em}
    \setlength{\tabcolsep}{1mm}
    \resizebox{\textwidth}{!}{
    \begin{tabular}{ccccccccc}
    \toprule
    \multirow{2}{*}{\shortstack{Strategy}} & 
    \multicolumn{2}{c}{\shortstack{NLG Metrics}} & 
    \multicolumn{5}{c}{\shortstack{CE Metrics}} &
    \multirow{2}{*}{\shortstack{Average F1$\triangle$}}
    \\
    &  BLEU-1 & ROUGE-L & Composition F1 & BI-RADS F1 & Finding F1 & Relation F1 & Average F1 & \\
    \midrule
    None &
    \textbf{53.30$\pm$0.32} & 62.46$\pm$0.25 & 41.22$\pm$1.90 & 24.29$\pm$0.53 & 74.57$\pm$0.37 & 39.58$\pm$0.26 & 44.92 & -  \\
    \midrule
    \cellcolor{gray!10}Cross-modal clinical knowledge integration  &
    53.20$\pm$0.36 & \textbf{62.53$\pm$0.26} & 45.61$\pm$2.03 & \textbf{25.17$\pm$0.56} & \textbf{75.33$\pm$0.39} & \textbf{40.13$\pm$0.26} & \textbf{46.56} & \textbf{+1.64}  \\
    \midrule
    Image-text retrieval  &
    52.81$\pm$0.38 & 61.69$\pm$0.27 & \textbf{45.76$\pm$2.11} & 24.50$\pm$0.52 & 74.55$\pm$0.40 & 40.03$\pm$0.27 & 46.21 & +1.29  \\
    \midrule
    Image-based retrieval &
    53.27$\pm$0.31 & 60.76$\pm$0.25 & 41.34$\pm$1.85 & 24.81$\pm$0.54 & 74.76$\pm$0.38 & 39.16$\pm$0.23 & 45.02 & +0.10  \\
    \bottomrule
    \end{tabular}}
    \label{tab:retrieval_ablation}
\end{table*}

(3) \textbf{Comparison with retrieval-based methods.}

To evaluate the effectiveness of the proposed cross-modal clinical knowledge integration framework, we compare our method with several retrieval-based strategies, as shown in Tab. \ref{tab:retrieval_ablation}. For a fair and concise comparison, only the \emph{patient-level feature} is adopted as auxiliary input. The “None” setting denotes the baseline without any auxiliary clinical information, where only image features are used for report generation.

We compare our method with two representative retrieval-based approaches: 1) \emph{image--text retrieval}, where a CLIP-style model is trained on the internal dataset using VersaMammo as the image encoder and Chinese-adapted BERT as the text encoder for 10 epochs, and the most semantically similar retrieved report is used as the patient-level feature; and 2) \emph{image-based retrieval}, where the average feature of the four views extracted by VersaMammo is used to retrieve the most visually similar patient image from the training set, and the corresponding report is adopted as auxiliary input.

From the results, several observations can be drawn. First, the proposed cross-modal clinical knowledge integration framework achieves the best overall performance, improving the CE average from 44.92\% to 46.56\% (+1.64\%), and obtaining the highest scores in BI-RADS F1 (25.17\%), Finding F1 (75.33\%), and Relation F1 (40.13\%). These results indicate that incorporating mammography-specific clinical priors into the generation process provides more effective diagnostic guidance for report generation.

Second, image--text retrieval also yields noticeable improvements (+1.29\% in CE average), particularly in Composition F1 (45.76\%), suggesting that semantically similar reports can provide useful high-level diagnostic context. However, it still underperforms compared to our method, indicating that semantic similarity alone is insufficient for accurately modeling clinically critical diagnostic attributes.

Finally, image-based retrieval brings only marginal gains (+0.10\%), showing that low-level visual similarity alone cannot effectively capture clinically meaningful diagnostic semantics, as visually similar mammograms may correspond to substantially different BI-RADS assessments and diagnostic conclusions.

Although the “None” setting achieves the highest BLEU-1 score, the difference is marginal. Moreover, BLEU-1 mainly measures surface-level lexical overlap and does not necessarily reflect clinical correctness or diagnostic consistency. In contrast, the proposed framework consistently achieves superior performance on clinically meaningful CE metrics, particularly BI-RADS F1 and Relation F1.

Overall, these results demonstrate that the proposed cross-modal clinical knowledge integration framework provides more effective clinical semantic modeling than conventional similarity-based approaches, leading to more accurate and clinically consistent report generation.

\begin{figure}
\centering
\includegraphics[width=0.8\textwidth]{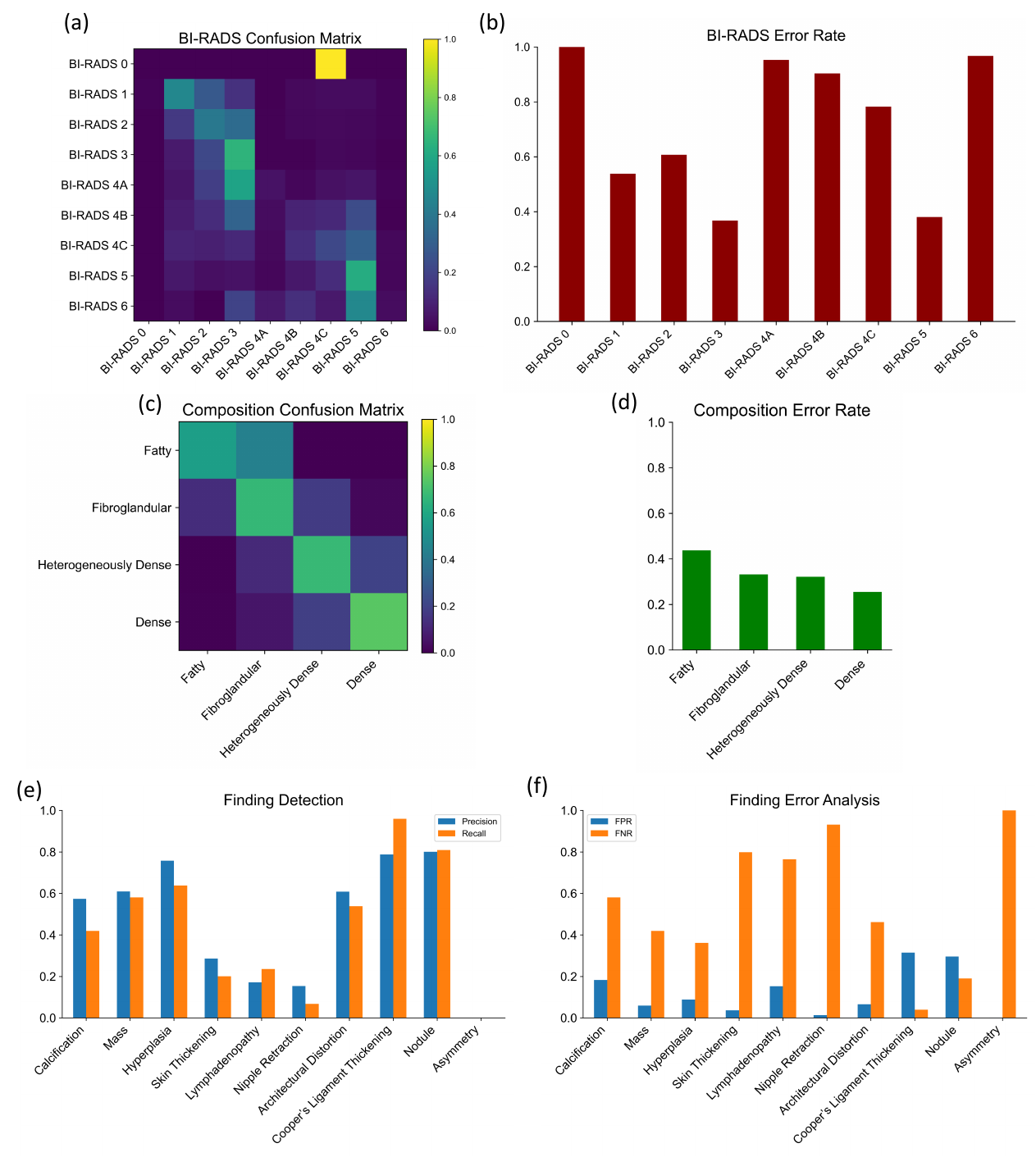}
\caption{\textbf{Error analysis of MammoRG predictions.}
(a) Confusion matrix of BI-RADS classification, showing the distribution of predicted categories against ground-truth labels.
(b) Error rates for each BI-RADS category, defined as the proportion of incorrect predictions within each class.
(c) Confusion matrix of breast composition classification.
(d) Error rates for each composition category.
(e-f) Finding-level analysis, including precision and recall, as well as false positives (FP) and false negatives (FN), reflecting detection accuracy, over-detection, and missed detection of different findings.}
\label{fig:analysis}
% \vspace{-1.5em}
\end{figure}

\subsubsection{Error Analysis}
The confusion matrix in Fig. \ref{fig:analysis} (a) reveals a clear tendency of the model to concentrate predictions around intermediate categories, particularly BI-RADS 3. While high-grade lesions such as BI-RADS 5 demonstrate relatively strong diagonal dominance (0.621), lower categories (BI-RADS 1–2) are frequently misclassified as higher-risk groups, especially BI-RADS 3. Notably, BI-RADS 4 subcategories (4A–4C) exhibit substantial overlap, with predictions often collapsing toward BI-RADS 3 or BI-RADS 5. This pattern suggests that the model struggles with fine-grained risk stratification, leading to systematic over-smoothing of diagnostic decisions. Fig. \ref{fig:analysis} (b) further quantifies these trends by presenting class-wise error rates. 

The composition confusion matrix in Fig. \ref{fig:analysis} (c) demonstrates relatively strong performance along the diagonal, particularly for Dense (0.745) and Heterogeneously Dense (0.679) categories. However, systematic confusion is observed between adjacent density levels, especially between Fatty and Fibroglandular, as well as between Heterogeneously Dense and Dense. This indicates that the model captures the general progression of breast density but struggles with precise categorization near class boundaries, which is consistent with the inherently continuous nature of density assessment. Fig. \ref{fig:analysis} (d) shows that classification accuracy improves monotonically with increasing density. The Dense category achieves the lowest error rate (0.255), while Fatty exhibits the highest (0.438). This suggests that high-density patterns are more visually distinctive and thus easier for the model to identify, whereas low-density breasts lack salient features and are more prone to misclassification. The results highlight a potential bias toward detecting more conspicuous tissue patterns.

Finding-level analysis in Fig. \ref{fig:analysis} (e-f) further reveals substantial variability in performance across different imaging findings. Overall, the model performs well on structurally prominent abnormalities, while exhibiting degraded performance on subtle or less conspicuous findings. Specifically, Nodule (Precision 0.801, Recall 0.809) and Cooper’s Ligament Thickening (Precision 0.788, Recall 0.960) demonstrate strong detection capability, suggesting that lesions with clear morphology and well-defined boundaries are more easily recognized. However, the relatively high false positive rates (FPR) for these categories (FPR 0.296 and 0.315, respectively) indicate a tendency toward over-prediction. This suggests that the model exhibits a pronounced hallucination tendency for these findings, potentially due to over-reliance on frequently occurring visual or textual patterns without sufficient grounding in actual image evidence. In contrast, several clinically important but visually subtle findings show significantly poorer performance. Asymmetry is entirely missed (Recall 0.000, False Negative Rate (FNR) 1.000), while Nipple Retraction (Recall 0.068, FNR 0.932) and Skin Thickening (Recall 0.201, FNR 0.799) also suffer from severe omission errors. These results indicate that the model struggles to capture fine-grained features and may additionally reflect the underrepresentation of these findings in the training data. Intermediate performance is observed for findings such as Calcification (Recall 0.419, FNR 0.581) and Architectural Distortion (Recall 0.538, FNR 0.462), indicating partial sensitivity but limited robustness.

Overall, the results highlight a systematic discrepancy between the model’s ability to recognize visually salient versus subtle abnormalities. While high recall on certain findings indicates strong sensitivity to prominent structural patterns, the coexistence of elevated false positive rates suggests limited specificity and a tendency toward pattern-driven over-prediction. At the same time, the model exhibits substantial failure in detecting low-saliency or underrepresented findings, leading to critical omission errors. This imbalance reflects a limitation in fine-grained clinical understanding, indicating that the model relies more on coarse visual cues than on precise feature discrimination. Improving sensitivity to subtle patterns and enhancing robustness to rare findings remain key directions for future improvement.

\section{Conclusion and Limitations}

In this study, we develop MammoRG, a two-stage framework for mammography report generation, together with MammoRGTool, a dedicated mammography report parsing tool for clinical efficacy evaluation. Our work addresses two important challenges in mammography report generation: the lack of mammography-specific clinical efficacy evaluation tools and the limited ability of existing models to capture mammography-specific clinical reasoning. Inspired by real-world clinical diagnostic workflows, the proposed framework explicitly incorporates mammography-specific clinical priors into the report generation process through cross-modal clinical knowledge integration. By modeling the mammography lexicon as a structured representation of BI-RADS-related diagnostic semantics and leveraging training reports as accumulated clinical experience, the framework enables better alignment between visual features and clinically relevant diagnostic semantics. In addition, the proposed term-aware tokenizer preserves mammography-specific medical semantics more effectively and alleviates the inefficiency caused by fragmented tokenization of professional terminology. Extensive experiments on four datasets demonstrate consistent improvements across both natural language generation and clinical efficacy metrics, particularly in BI-RADS assessment accuracy, highlighting the effectiveness and clinical applicability of the proposed framework.

Nevertheless, several limitations remain. First, the current model is trained only on Chinese mammography reports, and its generalizability to other languages and clinical environments remains unclear. Exploring cross-lingual and multilingual mammography report generation is an important direction for future work. Second, although this study focuses on mammography, extending the proposed framework to other imaging modalities would require modality-specific clinical knowledge representations, domain-specific terminology resources, and corresponding structured annotations. We hope this work can facilitate future research toward clinically reliable and knowledge-aware medical report generation systems.

\printcredits
\section*{Acknowledgement}
This work was supported by National Key R\&D Program of China (Project No. 2023YFE0204000), Hong Kong Innovation and Technology Commission (Project No. MHP/002/22) and Shenzhen Science and Technology Innovation Committee Fund (Project No. KCXFZ20230731094059008).

\section*{Data availability}
The VinDr-Mammo dataset is available and can be downloaded via the link \url{https://www.physionet.org/content/vindr-mammo/1.0.0/}. For the training dataset, the external 1 dataset, and the external 2 dataset are not publicly available due to patient privacy obligations, institutional review board requirements, and data use agreements. However, researchers interested in accessing deidentified data may submit a reasonable request directly to the corresponding authors, subject to obtaining the necessary ethical approvals and complying with institutional policies.

\section*{Code availability}
The trained model and source code can be accessed at \url{https://github.com/PiggyJerry/MammoRG}.
\bibliographystyle{cas-model2-names}

\bibliography{cas-refs}

@article{broeders2012impact,
  title={The impact of mammographic screening on breast cancer mortality in Europe: a review of observational studies},
  author={Broeders, Mireille and Moss, Sue and Nystr{\"o}m, Lennarth and Njor, Sisse and Jonsson, H{\aa}kan and Paap, Ellen and Massat, Nathalie and Duffy, Stephen and Lynge, Elsebeth and Paci, Eugenio},
  journal={Journal of Medical Screening},
  volume={19},
  number={1\_suppl},
  pages={14--25},
  year={2012},
  publisher={SAGE Publications Sage UK: London, England}
}

@article{morra2015breast,
  title={Breast cancer: computer-aided detection with digital breast tomosynthesis},
  author={Morra, Lia and Sacchetto, Daniela and Durando, Manuela and Agliozzo, Silvano and Carbonaro, Luca Alessandro and Delsanto, Silvia and Pesce, Barbara and Persano, Diego and Mariscotti, Giovanna and Marra, Vincenzo and others},
  journal={Radiology},
  volume={277},
  number={1},
  pages={56--63},
  year={2015},
  publisher={Radiological Society of North America}
}

@article{jalalian2013computer,
  title={Computer-aided detection/diagnosis of breast cancer in mammography and ultrasound: a review},
  author={Jalalian, Afsaneh and Mashohor, Syamsiah BT and Mahmud, Hajjah Rozi and Saripan, M Iqbal B and Ramli, Abdul Rahman B and Karasfi, Babak},
  journal={Clinical Imaging},
  volume={37},
  number={3},
  pages={420--426},
  year={2013},
  publisher={Elsevier}
}

@article{mammography2019diagnosis,
  title={Diagnosis and Staging of Breast Cancer: When and How to Use Mammography},
  author={Mammography, Contrast-Enhanced},
  journal={Diseases of the Chest, Breast, Heart and Vessels 2019-2022: Diagnostic and Interventional Imaging},
  pages={155},
  year={2019},
  publisher={Springer}
}

@inproceedings{abbey2020sequential,
  title={Sequential reading effects in Dutch screening mammography},
  author={Abbey, Craig K and Webster, Michael A and Geertse, Tanya and van der Waal, Danielle and Tetteroo, Eric and Pijnappel, Ruud and Broeders, Mireille JM and Sechopoulos, Ioannis},
  booktitle={Medical Imaging 2020: Image Perception, Observer Performance, and Technology Assessment},
  volume={11316},
  pages={66--70},
  year={2020},
  organization={SPIE}
}

@article{zambrano2025clinically,
  title={A clinically accessible small multimodal radiology model and evaluation metric for chest X-ray findings},
  author={Zambrano Chaves, Juan Manuel and Huang, Shih-Cheng and Xu, Yanbo and Xu, Hanwen and Usuyama, Naoto and Zhang, Sheng and Wang, Fei and Xie, Yujia and Khademi, Mahmoud and Yang, Ziyi and others},
  journal={Nature Communications},
  volume={16},
  number={1},
  pages={3108},
  year={2025},
  publisher={Nature Publishing Group UK London}
}

@article{li2025towards,
  title={Towards a holistic framework for multimodal LLM in 3D brain CT radiology report generation},
  author={Li, Cheng-Yi and Chang, Kao-Jung and Yang, Cheng-Fu and Wu, Hsin-Yu and Chen, Wenting and Bansal, Hritik and Chen, Ling and Yang, Yi-Ping and Chen, Yu-Chun and Chen, Shih-Pin and others},
  journal={Nature Communications},
  volume={16},
  number={1},
  pages={2258},
  year={2025},
  publisher={Nature Publishing Group UK London}
}

@article{lei2025interpretable,
  title={Interpretable Brain MRI Report Generation Anchored by Lesion Topography},
  author={Lei, Jiayu and Zhang, Xiaoman and Wu, Chaoyi and Dai, Lisong and Zhang, Ya and Zhang, Yanyong and Wang, Yanfeng and Xie, Weidi and Li, Yuehua},
  journal={IEEE Journal of Biomedical and Health Informatics},
  year={2025},
  publisher={IEEE}
}

@article{tran2025generating,
  title={Generating dermatopathology reports from gigapixel whole slide images with HistoGPT},
  author={Tran, Manuel and Schmidle, Paul and Guo, Ruifeng Ray and Wagner, Sophia J and Koch, Valentin and Lupperger, Valerio and Novotny, Brenna and Murphree, Dennis H and Hardway, Heather D and D’Amato, Marina and others},
  journal={Nature Communications},
  volume={16},
  number={1},
  pages={4886},
  year={2025},
  publisher={Nature Publishing Group UK London}
}

@article{liberman2002breast,
  title={Breast imaging reporting and data system (BI-RADS)},
  author={Liberman, Laura and Menell, Jennifer H},
  journal={Radiologic Clinics},
  volume={40},
  number={3},
  pages={409--430},
  year={2002},
  publisher={Elsevier}
}

@inproceedings{yalunin2021generating,
  title={Generating mammography reports from multi-view mammograms with bert},
  author={Yalunin, Alexander and Sokolova, Elena and Burenko, Ilya and Ponomarchuk, Alexander and Puchkova, Olga and Umerenkov, Dmitriy},
  booktitle={Findings of the Association for Computational Linguistics: EMNLP 2021},
  pages={153--162},
  year={2021}
}

@article{de2025convnext,
  title={A ConvNeXt-Transformer Approach for Automated Conclusion Generation from Mammography},
  author={de Avila Armenta, Eduardo and Bosques-Palomo, Beatriz and {\'A}lez, Gerardo A Fumagal-Gonz and Monsivais-Molina, Mario A and Garza-Abdala, Jorge A and Hussain, Sadam and Vela-Jarquin, Daniel and Cardona-Huerta, Servando and {\~N}o-Avalos, Daly B Avenda and {\~N}a, Jose G Tamez-Pe},
  journal={Authorea Preprints},
  year={2025},
  publisher={Authorea}
}

@article{cao2025mammovlm,
  title={MammoVLM: A generative large vision--language model for mammography-related diagnostic assistance},
  author={Cao, Zhenjie and Deng, Zhuo and Ma, Jie and Hu, Jintao and Ma, Lan},
  journal={Information Fusion},
  volume={118},
  pages={102998},
  year={2025},
  publisher={Elsevier}
}

@inproceedings{smit2020combining,
  title={Combining automatic labelers and expert annotations for accurate radiology report labeling using BERT},
  author={Smit, Akshay and Jain, Saahil and Rajpurkar, Pranav and Pareek, Anuj and Ng, Andrew Y and Lungren, Matthew},
  booktitle={Proceedings of the 2020 Conference on Empirical Methods in Natural Language Processing (EMNLP)},
  pages={1500--1519},
  year={2020}
}

@article{yang2025qwen3,
  title={Qwen3 technical report},
  author={Yang, An and Li, Anfeng and Yang, Baosong and Zhang, Beichen and Hui, Binyuan and Zheng, Bo and Yu, Bowen and Gao, Chang and Huang, Chengen and Lv, Chenxu and others},
  journal={arXiv preprint arXiv:2505.09388},
  year={2025}
}

@inproceedings{jin2024promptmrg,
  title={Promptmrg: Diagnosis-driven prompts for medical report generation},
  author={Jin, Haibo and Che, Haoxuan and Lin, Yi and Chen, Hao},
  booktitle={Proceedings of the AAAI Conference on Artificial Intelligence},
  volume={38},
  number={3},
  pages={2607--2615},
  year={2024}
}

@article{liu2021auto,
  title={Auto-encoding knowledge graph for unsupervised medical report generation},
  author={Liu, Fenglin and You, Chenyu and Wu, Xian and Ge, Shen and Sun, Xu and others},
  journal={Advances in Neural Information Processing Systems},
  volume={34},
  pages={16266--16279},
  year={2021}
}

@inproceedings{ranjit2023retrieval,
  title={Retrieval augmented chest x-ray report generation using openai gpt models},
  author={Ranjit, Mercy and Ganapathy, Gopinath and Manuel, Ranjit and Ganu, Tanuja},
  booktitle={Machine Learning for Healthcare Conference},
  pages={650--666},
  year={2023},
  organization={PMLR}
}

@article{wang2020unifying,
  title={Unifying relational sentence generation and retrieval for medical image report composition},
  author={Wang, Fuyu and Liang, Xiaodan and Xu, Lin and Lin, Liang},
  journal={IEEE Transactions on Cybernetics},
  volume={52},
  number={6},
  pages={5015--5025},
  year={2020},
  publisher={IEEE}
}

@inproceedings{nishino2022factual,
  title={Factual accuracy is not enough: Planning consistent description order for radiology report generation},
  author={Nishino, Toru and Miura, Yasuhide and Taniguchi, Tomoki and Ohkuma, Tomoko and Suzuki, Yuki and Kido, Shoji and Tomiyama, Noriyuki},
  booktitle={Proceedings of the 2022 Conference on Empirical Methods in Natural Language Processing},
  pages={7123--7138},
  year={2022}
}

@article{najdenkoska2022uncertainty,
  title={Uncertainty-aware report generation for chest X-rays by variational topic inference},
  author={Najdenkoska, Ivona and Zhen, Xiantong and Worring, Marcel and Shao, Ling},
  journal={Medical Image Analysis},
  volume={82},
  pages={102603},
  year={2022},
  publisher={Elsevier}
}

@inproceedings{devlin2019bert,
  title={Bert: Pre-training of deep bidirectional transformers for language understanding},
  author={Devlin, Jacob and Chang, Ming-Wei and Lee, Kenton and Toutanova, Kristina},
  booktitle={Proceedings of the 2019 Conference of the North American Chapter of the Association for Computational Linguistics: Human Language Technologies, volume 1 (long and short papers)},
  pages={4171--4186},
  year={2019}
}

@article{moon2022multi,
  title={Multi-modal understanding and generation for medical images and text via vision-language pre-training},
  author={Moon, Jong Hak and Lee, Hyungyung and Shin, Woncheol and Kim, Young-Hak and Choi, Edward},
  journal={IEEE Journal of Biomedical and Health Informatics},
  volume={26},
  number={12},
  pages={6070--6080},
  year={2022},
  publisher={IEEE}
}

@article{selivanov2023medical,
  title={Medical image captioning via generative pretrained transformers},
  author={Selivanov, Alexander and Rogov, Oleg Y and Chesakov, Daniil and Shelmanov, Artem and Fedulova, Irina and Dylov, Dmitry V},
  journal={Scientific Reports},
  volume={13},
  number={1},
  pages={4171},
  year={2023},
  publisher={Nature Publishing Group UK London}
}

@inproceedings{irvin2019chexpert,
  title={Chexpert: A large chest radiograph dataset with uncertainty labels and expert comparison},
  author={Irvin, Jeremy and Rajpurkar, Pranav and Ko, Michael and Yu, Yifan and Ciurea-Ilcus, Silviana and Chute, Chris and Marklund, Henrik and Haghgoo, Behzad and Ball, Robyn and Shpanskaya, Katie and others},
  booktitle={Proceedings of the AAAI Conference on Artificial Intelligence},
  volume={33},
  number={01},
  pages={590--597},
  year={2019}
}

@article{yan2023attributed,
  title={Attributed abnormality graph embedding for clinically accurate x-ray report generation},
  author={Yan, Sixing and Cheung, William K and Chiu, Keith and Tong, Terence M and Cheung, Ka Chun and See, Simon},
  journal={IEEE Transactions on Medical Imaging},
  volume={42},
  number={8},
  pages={2211--2222},
  year={2023},
  publisher={IEEE}
}

@inproceedings{NEURIPSDATASETSANDBENCHMARKS2021_c8ffe9a5,
 author = {Jain, Saahil and Agrawal, Ashwin and Saporta, Adriel and Truong, Steven and Duong, Du Nguyen Duong Nguyen and Bui, Tan and Chambon, Pierre and Zhang, Yuhao and Lungren, Matthew and Ng, Andrew and Langlotz, Curtis and Rajpurkar, Pranav and Rajpurkar, Pranav},
 booktitle = {Proceedings of the Neural Information Processing Systems Track on Datasets and Benchmarks},
 editor = {J. Vanschoren and S. Yeung},
 pages = {},
 publisher = {Curran},
 title = {RadGraph: Extracting Clinical Entities and Relations from Radiology Reports},
 url = {https://datasets-benchmarks-proceedings.neurips.cc/paper_files/paper/2021/file/c8ffe9a587b126f152ed3d89a146b445-Paper-round1.pdf},
 volume = {1},
 year = {2021}
}

@inproceedings{shang2022onerel,
  title={Onerel: Joint entity and relation extraction with one module in one step},
  author={Shang, Yu-Ming and Huang, Heyan and Mao, Xianling},
  booktitle={Proceedings of the AAAI Conference on Artificial Intelligence},
  volume={36},
  number={10},
  pages={11285--11293},
  year={2022}
}

@article{radford2019language,
  title={Language models are unsupervised multitask learners},
  author={Radford, Alec and Wu, Jeffrey and Child, Rewon and Luan, David and Amodei, Dario and Sutskever, Ilya and others},
  journal={OpenAI blog},
  volume={1},
  number={8},
  pages={9},
  year={2019}
}

@article{nguyen2023vindr,
  title={VinDr-Mammo: A large-scale benchmark dataset for computer-aided diagnosis in full-field digital mammography},
  author={Nguyen, Hieu T and Nguyen, Ha Q and Pham, Hieu H and Lam, Khanh and Le, Linh T and Dao, Minh and Vu, Van},
  journal={Scientific Data},
  volume={10},
  number={1},
  pages={277},
  year={2023},
  publisher={Nature Publishing Group UK London}
}

@inproceedings{papineni2002bleu,
  title={Bleu: a method for automatic evaluation of machine translation},
  author={Papineni, Kishore and Roukos, Salim and Ward, Todd and Zhu, Wei-Jing},
  booktitle={Proceedings of the 40th Annual Meeting of the Association for Computational Linguistics},
  pages={311--318},
  year={2002}
}

@inproceedings{lin2004rouge,
  title={Rouge: A package for automatic evaluation of summaries},
  author={Lin, Chin-Yew},
  booktitle={Text Summarization Branches Out},
  pages={74--81},
  year={2004}
}

@article{huang2025versatile,
  title={A Versatile Foundation Model for AI-enabled Mammogram Interpretation},
  author={Huang, Fuxiang and Zhu, Jiayi and Yu, Yunfang and Xie, Yu and Guo, Yuan and Kong, Qingcong and Wu, Mingxiang and Jiang, Xinrui and Yang, Shu and Ma, Jiabo and others},
  journal={arXiv preprint arXiv:2509.20271},
  year={2025}
}

@article{zhu2025benchmark,
  title={A Benchmark for Breast Cancer Screening and Diagnosis in Mammogram Visual Question Answering},
  author={Zhu, Jiayi and Huang, Fuxiang and Luo, Qiong and Chen, Hao},
  journal={Nature Communications},
  year={2025},
  publisher={Nature Publishing Group UK London}
}

@article{hu2022lora,
  title={Lora: Low-rank adaptation of large language models.},
  author={Hu, Edward J and Shen, Yelong and Wallis, Phillip and Allen-Zhu, Zeyuan and Li, Yuanzhi and Wang, Shean and Wang, Lu and Chen, Weizhu and others},
  journal={The International Conference on Learning Representations},
  volume={1},
  number={2},
  pages={3},
  year={2022}
}

@article{Qwen3-VL,
      title={Qwen3-VL Technical Report}, 
      author={Shuai Bai and Yuxuan Cai and Ruizhe Chen and Keqin Chen and Xionghui Chen and Zesen Cheng and Lianghao Deng and Wei Ding and Chang Gao and Chunjiang Ge and Wenbin Ge and Zhifang Guo and Qidong Huang and Jie Huang and Fei Huang and Binyuan Hui and Shutong Jiang and Zhaohai Li and Mingsheng Li and Mei Li and Kaixin Li and Zicheng Lin and Junyang Lin and Xuejing Liu and Jiawei Liu and Chenglong Liu and Yang Liu and Dayiheng Liu and Shixuan Liu and Dunjie Lu and Ruilin Luo and Chenxu Lv and Rui Men and Lingchen Meng and Xuancheng Ren and Xingzhang Ren and Sibo Song and Yuchong Sun and Jun Tang and Jianhong Tu and Jianqiang Wan and Peng Wang and Pengfei Wang and Qiuyue Wang and Yuxuan Wang and Tianbao Xie and Yiheng Xu and Haiyang Xu and Jin Xu and Zhibo Yang and Mingkun Yang and Jianxin Yang and An Yang and Bowen Yu and Fei Zhang and Hang Zhang and Xi Zhang and Bo Zheng and Humen Zhong and Jingren Zhou and Fan Zhou and Jing Zhou and Yuanzhi Zhu and Ke Zhu},
	  journal={arXiv preprint arXiv:2511.21631},
      year={2025}
}

@article{wang2025internvl3,
  title={Internvl3. 5: Advancing open-source multimodal models in versatility, reasoning, and efficiency},
  author={Wang, Weiyun and Gao, Zhangwei and Gu, Lixin and Pu, Hengjun and Cui, Long and Wei, Xingguang and Liu, Zhaoyang and Jing, Linglin and Ye, Shenglong and Shao, Jie and others},
  journal={arXiv preprint arXiv:2508.18265},
  year={2025}
}

@article{xu2025lingshu,
  title={Lingshu: A Generalist Foundation Model for Unified Multimodal Medical Understanding and Reasoning},
  author={Xu, Weiwen and Chan, Hou Pong and Li, Long and Aljunied, Mahani and Yuan, Ruifeng and Wang, Jianyu and Xiao, Chenghao and Chen, Guizhen and Liu, Chaoqun and Li, Zhaodonghui and others},
  journal={arXiv preprint arXiv:2506.07044},
  year={2025}
}

@inproceedings{chen-emnlp-2020-r2gen,
    title = "Generating Radiology Reports via Memory-driven Transformer",
    author = "Chen, Zhihong and
      Song, Yan  and
      Chang, Tsung-Hui and
      Wan, Xiang",
    booktitle = "Proceedings of the 2020 Conference on Empirical Methods in Natural Language Processing",
    month = nov,
    year = "2020",
}

@article{yang2023radiology,
  title={Radiology report generation with a learned knowledge base and multi-modal alignment},
  author={Yang, Shuxin and Wu, Xian and Ge, Shen and Zheng, Zhuozhao and Zhou, S Kevin and Xiao, Li},
  journal={Medical Image Analysis},
  volume={86},
  pages={102798},
  year={2023},
  publisher={Elsevier}
}

@article{spak2017bi,
  title={BI-RADS{\textregistered} fifth edition: A summary of changes},
  author={Spak, David Allen and Plaxco, JS and Santiago, L and Dryden, MJ and Dogan, BE},
  journal={Diagnostic and Interventional Imaging},
  volume={98},
  number={3},
  pages={179--190},
  year={2017},
  publisher={Elsevier}
}

@article{singh2025openai,
  title={Openai gpt-5 system card},
  author={Singh, Aaditya and Fry, Adam and Perelman, Adam and Tart, Adam and Ganesh, Adi and El-Kishky, Ahmed and McLaughlin, Aidan and Low, Aiden and Ostrow, AJ and Ananthram, Akhila and others},
  journal={arXiv preprint arXiv:2601.03267},
  year={2025}
}

\end{document}